# Predicting the spread of COVID-19 in Delhi, India using Deep Residual Recurrent Neural Networks.

Shashank Reddy Vadyala
Department of Computational Analysis and Modeling, Louisiana Tech University, Ruston, Louisiana, United States.
Srv009@email.latech.edu

Sai Nethra Betgeri
Department of Computational Analysis and Modeling, Louisiana Tech University, Ruston, Louisiana, United States.
Snb023@email.latech.edu

Detecting the spread of coronavirus (COVID-19) will go a long way toward reducing human and economic loss. Unfortunately, existing Epidemiological models used for COVID-19 prediction models are too slow and fail to capture the COVID-19 development in detail. This research uses Partial Differential Equations (PDEs) to improve the processing speed and accuracy of forecasting of COVID -19 governed by SEIRD model equations. The dynamics of COVID-19 were extracted using Convolutional Neural Networks (CNNs) and Deep Residual Recurrent Neural Networks (DR-RNNs) from data simulated using PDEs. The DR-RNNs accuracy is measured using Mean Squared Error (MSE). The DR-RNNs COVID-19 prediction model has been shown to have accurate COVID-19 predictions. In addition, we concluded that DR-RNNs can significantly advance the ability to support decision-making in real-time COVID-19 prediction.

## 1.Introduction

The COVID-19 epidemic is undoubtedly the most severe public health threat since the Spanish flu outbreaks of 1918 and 1919(Wang et al., 2020). Following a dramatic increase in COVID-19 infections, many nations, including the United States of America (USA) and India, declared states of emergency. The World Health Organization (WHO) said the outbreak has affected 25,227,970 people and killed 278,751 people in India as of May 17, 2021, causing hospitals in India to overflow. In addition, COVID-19 has created mayhem in the financial sector, resulting in the S&P 500's worst trading day since 1945(Bhadra et al., 2021). On December 8, 2019, COVID-19, caused by the Severe Acute Respiratory Syndrome COVID-19, was discovered in Wuhan, China("Clinical characteristics of refractory COVID-19 pneumonia in Wuhan, China | Clinical Infectious Diseases | Oxford Academic," n.d.). Most of the first patients have been exposed to the nearby Huanan South China seafood market, which offers several wild animals, implying that the zoonotic coronavirus breached the animal-human boundary at this wet market. COVID-19, namely the Severe Acute Respiratory Syndrome and the Middle East Respiratory Syndrome (MERSV), have caused a couple of significant epidemics in the last twenty years. (Petrosillo et al., 2020).

COVID-19 drastic impact on our social lives and the economy has piqued scientists' interest in this novel virus. However, multiple significant concerns regarding the pandemic remain unanswered at this period (Maggi et al., 2020). Aside from pathology, microbiology, and bioinformatics, the COVID-19 epidemic has sparked interest in epidemiology and statistics. Time series analysis, machine learning models, and forecasting models are of particular interest(Ahmed et al., 2010). Critical risk assessment and coordination countermeasures should be taken to aid an accurate forecast of future events. Forecasting COVID-19 is critical for understanding the estimation of virus transmission characteristics such as the basic reproduction number, incubation period, and infectious period(Vadyala et al., 2020). In a real scenario, those parameters are not easy to estimate. Accurate physics-based models for forecasting COVID-19 require solving complex partial differential equations (PDEs).

PDEs have played a critical role in providing detailed and reliable models for various scientific phenomena, especially physics and engineering(Barbu, 2020). PDEs control physical phenomena such as the Navier–Stokes equation in fluid, aerodynamics, Fourier's heat conduction equation, and Schrödinger's equation in quantum mechanics. These models were first discovered using skills in philosophy, statistical modeling, and observational evidence. Equations that seem to model seemingly disparate physical processes are identical in that they are made up of specific, commonly important mathematical components. Physical behaviors placed on the observable data by various parts of the model may aid experts in identifying the model using prior theory and data information(Carleo et al., 2019). We can now view massive volumes of data from tests and simulations thanks to recent advances in data acquisition, storage, and computing tools. This makes it possible to derive information from raw data using data-driven approaches. DR-RNNs has revolutionized many areas, such as machine vision, in recent decades(Vadyala and Betgeri, 2021). When extended to scientific evidence, they can help solve and explain physical problems.

DR-RNNs includes the physics of the underlying problem in the loss function(Kani and Elsheikh, 2017). The governing PDEs are used to directly calculate the loss function of DR-RNNs, which is minimized during training. The residuals at collocation points, the weighted residuals obtained by the Galerkin-Method, or the energy functional of an Euler-Lagrange differential equation all contribute to the loss function(Nabian and Meidani, 2019). Thus, for training DR-RNNs, there is no need to generate labeled training data in advance. Instead, in the problem domain, the input data for DR-RNNs points are sampled. Distance between the approximated solution of the PDEs and measured values of the solution is augmented by the loss function. Then one or more coefficients of the PDEs can be included as unknowns during training. By using this way, an inverse problem is solved. Solving inverse problems with DR-RNNs may lead to a significant speedup compared to conventional methods(Vadyala and Betgeri, 2021). For example, DR-RNNs solve eigenvalue problems when the loss function of a neural network is related to the Rayleigh-Ritz coefficient(Kovacs et al., 2021). Physics-based machine learning models that could offer faster and potentially more accurate solutions are essential to replace existing numerical epidemiological models.

To conduct this research, we took the following approach: We start with a SIERD model and use analytics to derive a differential equation of infection rate, which provides information on individual infection percentages in the region. The infection rate is then calculated using the heatmap snapshots for the region-infected community. This process is carried out using Machine Learning techniques, specifically DR-RNNs. The latter was pre-trained on simulated SIERD data before being trained on each region's recorded infected data. Finally, we validate the resulting SIERD model against the region's data until we know the infection rate. The following is a breakdown of the paper's structure. Section 2 talks about the methodology of the study introduce the procedure to produce data for DR-RNNs. The performance of DR-RNNs is evaluated and discussed in Section 3. Section 4 summarizes the limitation of the study, and section 5 concludes.

## 2. Methods:

The DR-RNNs are trained on large quantities of data simulated by the SEIRD model PDEs. Finally, the computational efficiency of the DR-RNNs is investigated. Our proposed model DR-RNNs uses the snapshots of COVID-19 data of Delhi, India, from January 1 to April 30, 2021, to forecast the potential growth of the COVID-19 for the next two weeks. In addition, data were obtained from the arcgis.

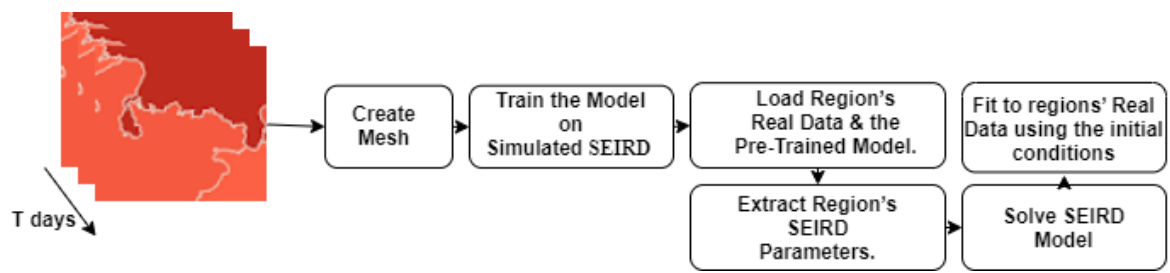

Figure.1. Overview of DR-RNNs for SEIRD model

In this Figure.1, we describe the overview of DR-RNNs through a flowchart. We note that after the extraction of the infection rate from the data, we use the SEIRD model to investigate the parameter further. Gmsh is used to build the mesh, which is uniformly optimized as the simulation progresses. After being developed in one level, the mesh has a minimum spatial resolution of around 1 kilometer.

## 2.1. Convolutional Neural Network

Yann André LeCun proposed Convolutional Neural Networks (CNNs) in 1998(Bengio and Lecun, 1997). The net has eight layers with weights, as shown in Figure.2, with the first five being convolutional and the last three being connected. The previous completely connected layer's output is fed into a 1000-way softmax, distributing the 1000 class labels. Since the data representation to our neural network is an image (2D).

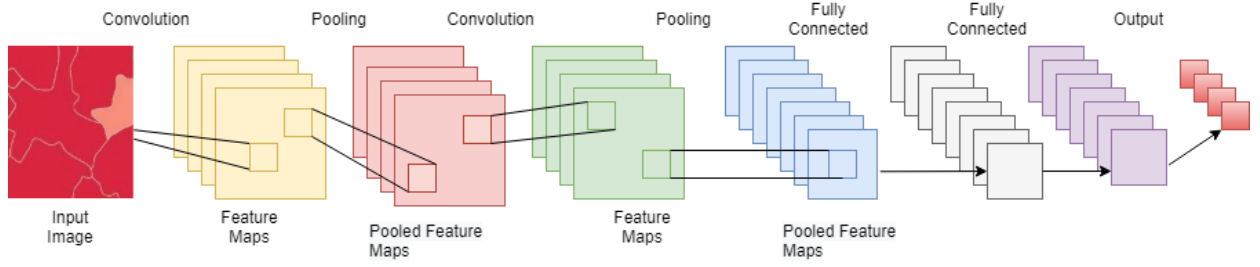

Figure. 2. CNNs overview

## 2.2. LSTM network

The Recurrent Neural Networks (RNN) architecture can be written as:

$$h_t = tanh(W[h_{t-1}, z_t] + b) \tag{1}$$

$$x_t = Vh_t + c \tag{2}$$

$$L(\theta) = \frac{1}{TSN}\sum_{S=1}^{S}\sum_{n}^{N}\sum_{t=1}^{T} |\hat{y}_t - y_t| \tag{3}$$

Where $z_t$ is the input at time $t$; $h_t$ is the hidden state at time $t$; $x_t$ is the output at time $t$; $b$ and $c$ are the bias vectors of the RNN; $W$ and $V$ are the weight matrices of the RNN are shown in Eq. (2). The current time step's input and the previous time step's hidden state are used in calculating the current hidden state., which acts as the RNN's memory, according to Eq. (1). For predicting the responses of dynamical systems, the loss function is shown in Eq. (3). Where $\hat{y}_t$ and $y_t$ are the predicted and actual system states at given time $t$. The number of time measures, functions, and training samples is denoted by $T$, $N$, and $S$, respectively. Back-propagation through time is used to change the values of RNN parameters during preparation (Plaut and And Others, 1986). However, it was reported that the standard RNN architecture has difficulties in learning long-term dependencies due to the vanishing or exploding gradient problem(Schaefer et al., 2008). Therefore, Gated RNN designs such as the LSTM and the gated recurrent unit were developed to overcome this issue(Sherstinsky, 2020; Tjandra et al., 2016). RNNs that can learn long-term dependencies are known as LSTM networks. An LSTM unit contains three gates, i.e., input gate, forget gate, and output gate. The LSTM architecture can be expressed in the following Eq.(4-8).

$$f_t = \sigma_s(W_f[h_{t-1}, z_t] + b_f) \tag{4}$$

$$i_t = \sigma_s(W_i[h_{t-1}, z_t] + b_i) \tag{5}$$

$$o_t = \sigma_s(W_o[h_{t-1}, z_t] + b_o) \tag{6}$$

$$c_t = f_t * c_{t-1} + i_t * \sigma_h((W_c[h_{t-1}, z_t] + b_c) \tag{7}$$

$$h_t = o_t * \sigma_h(c_t) \tag{8}$$

where $f_t$ is the forget gate vector; it is the input gate vector; $o_t$ is the output gate vector; $h_{t-1}$ is the hidden state vector; $c_{t-1}$ is the cell state vector, $W_c$, $W_i$, $W_o$ and $W_f$ are the weight matrices; $b_c, b_o, b_i$ and $b_f$ are the bias vectors; the operator $*$ denotes element-wise multiplication; $\sigma_s$ represents the sigmoid function; and $\sigma_h$ represents the hyperbolic tangent function.

### 2.3. Deep Residual Recurrent Neural Networks (DR-RNNs)

The physics of dynamical systems is expressed in their governing equations, which can be written in a general form as:

$$dy/dt = f(t, y) \tag{9}$$

Where y is the dynamical system's state variable, Eq. (9) can be solved analytically or numerically to obtain the dynamical system's answer. For example, using the implicit Euler form, the device state at time instant $t + 1$ can be obtained as:

$$y_{t+1} = y_t + h \times f(t+1, y_{t+1}) \tag{10}$$

where $h$ is the time step size. From Eq. (10), a residual function can be formulated as:

$$r_{t+1} = y_{t+1} - y_t + h \times f(t+1, y_{t+1}) \tag{11}$$

By stacking I network layers as shown in Eq.(12), the DR-RNNs architecture is meant to iteratively reduce the residual function provided in Eq. (11).

$$y_{t+1}^i = y_{t+1}^{i-1} - W * tanh(Ur_{t+1}^i), \text{ for i=1} \tag{12}$$

$$y_{t+1}^i = y_{t+1}^{i-1} - \frac{\eta_k}{\sqrt{G_k+\alpha}} r_{t+1}^i, \text{ for i>1} \tag{13}$$

where i is the layer number; $r_{t+1}^i$ is the residual at time instant $t + 1$ in the ith layer as shown in Eq.(7-8); $W, U$, and $\eta$ are the weight parameters of the DR-RNN(Kani and Elsheikh, 2017); To avoid division by zero $\alpha$ is a used which is a small number; and $H_i$ is determined as the residual's exponentially decreasing squared norm

$$H_i = \gamma||r_{t+1}||^2 + \beta H_{i-1} \tag{14}$$

where $\beta$ and $\gamma$ are the fraction factors, and their values are set as 0.9 and 0.1, respectively. The training objective of the DR-RNNs is to find a set of parameters that minimizes the residual function defined in Eq. (11). The DR-RNNs can be thought of as a numerical integrator, which is, to a great extent, like performing implicit integration i.e.making the residual zero by solving Eq. (10). Thus, DR-RNNs are learning to perform implicit integration. However, the DR-RNNs are explicit in time with a constant computational cost at each step, unlike implicit integration methods. To minimize the loss function given in DR-RNNs, the Adam optimizer (Bock and Weiß, 2019) is used, which was created using the TensorFlow machine learning system in Eq.(3).

## 2.4.Modeling and integration of COVID-19

In this section, the COVID-19 is first briefly reviewed, and then the integration of the equations of SIERD into the DR-RNNs is introduced. The development of short-term prediction models for forecasting the number of possible cases, aided by computational simulation of the virus's dynamics. To avoid deaths and cure patients, strategic planning should be implemented in the public health industry. Using a computational model's virus transmission can be forecasted. Here, we work with a spatio-temporal SEIRD model, presented in (Viguerie et al., 2021), and given in Eq.(15-19).

$$\frac{\partial s}{\partial t} + \varphi_i \left(1 - \frac{A_e}{n_p}\right) si + \varphi_e \left(1 - \frac{A_e}{n_p}\right) se - \nabla \cdot (n_p v_s \nabla s) = 0 \tag{15}$$

$$\frac{\partial e}{\partial t} - \varphi_i \left(1 - \frac{A_e}{n_p}\right) si - \varphi_i \left(1 - \frac{A_e}{n_p}\right) se + (\alpha + \gamma_e)e - \nabla \cdot (n_p v_e \nabla e) = 0 \tag{16}$$

$$\frac{\partial i}{\partial t} - \alpha e + (\gamma_i i + \delta)i - \nabla \cdot (n_p v_r \nabla i) = 0 \tag{17}$$

$$\frac{\partial i}{\partial t} - \gamma_e e - \gamma_i i - \nabla \cdot (n_p v_r \nabla r) = 0 \tag{18}$$

$$\frac{\partial d}{\partial t} - \delta i = 0 \tag{19}$$

where the densities of the susceptible, exposed, infectious, recovered, and deceased populations are denoted by $s(x; t)$, $e(x; t)$, $i(x; t)$, $r(x; t)$, and $d(x; t)$ respectively. The total living population is expressed by $n_p$ Which is the number of all compartments except $d(x; t)$. we considered the tendency of outbreaks to cluster around large populations, $\varphi_i$ and $\varphi_e$ denote the transmission rates between symptomatic and susceptible individuals and asymptomatic and susceptible individuals, respectively (units $\text{days}^{-1}$), $\alpha$ denotes the incubation period (units $\text{days}^{-1}$), $\gamma_e$ corresponds to the asymptomatic recovery rate (units $\text{days}^{-1}$), $\gamma_i$ the symptomatic recovery rate (units $\text{days}^{-1}$), represents the mortality rate (units $\text{days}^{-1}$), and $v_s, v_e, v_i, v_r$ are the diffusion parameters of the different population groups as denoted by the sub-scripted letters (units $km^2$ $persons^{-1}$ $days^{-1}$). Note that all these parameters can be considered time and space-dependent.

## 2.5. Learning virus dynamics

The snapshots can now be assembled into a snapshot matrix for the training DR-RNNs. Once we have DR-RNNs trained on the infected data of the region, we may use it to extract the presence and persistence of the social distancing measures typified through the function. The reference target mesh is the uniformly refined fixed mesh considered in the early stages of the simulation, presenting 56558 nodes and 78340 elements. The simulation considers a time step size of $\Delta t = 0:25$ days for the numerical integration and $\Delta t_0 = \Delta t = 0:25$ days for the observations. We then run fixed mesh simulations. The simulation of the SIERD model, obtained using the Runge-Kutta method. The accuracy of the learned parameters by DR-RNNs is validated using the MSE with respect to the exact solution. We start by presenting the results for the daily learned parameters, followed by the associated reproduction numbers. Then, predictions of infectious cases are provided using the initial conditions as shown in Figure.3. The details of the training are explained below.

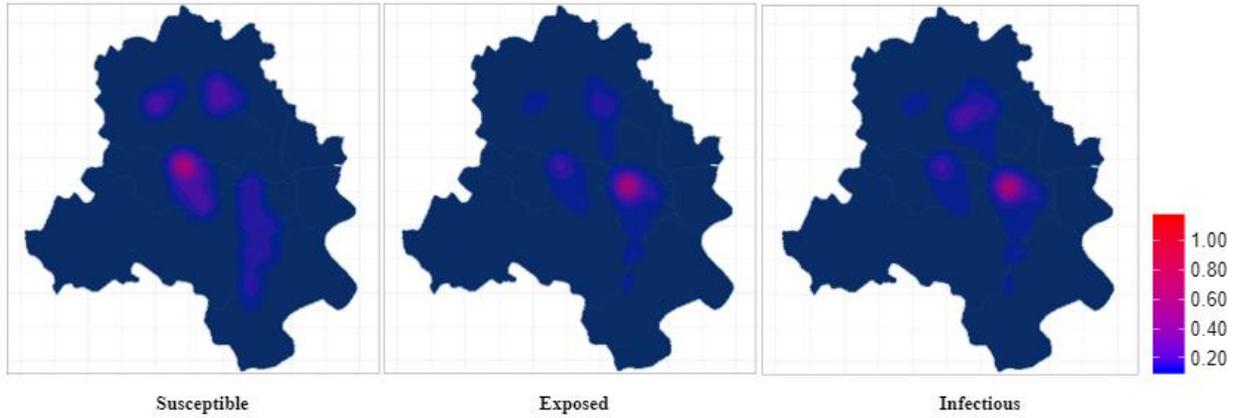

Figure.3. Initial conditions (January 1, 2021) in Delhi, India

The model is trained using a custom training loop by minimizing the mean squared error loss for data ($MSE_u$) as shown in Eq.(20).

$$MSE_u = \frac{1}{N}\sum_{i=1}^{N} |y_t - \hat{y}_t|^2 \qquad (20)$$

where $\{x_t, \hat{x}_t\}_{t=1}^{N}$ denote the set of the reported, $x_t$ real reported cases and $\hat{x}_t$ represents predicted cases and the mean squared error loss ($MSE_s$) represents the residuals obtained from the SIERD model Eq. (15-19) by subtracting the right side from the left side.

$$MSE_L = \omega_u MSE_u + \omega_s MSE_s \qquad (21)$$

The loss function of DR-RNNs is made of two terms: $MSE_u$ and $MSE_s$ as shown in Eq.(21). Which are defined by the last part of the above equation. The parameters $\{\omega_u, \omega_s\}$ denote the weight coefficients in the loss function that can balance the optimization effort between learning the data and satisfying the SEIRD model PDEs. We load the pre-trained model and allow all its weights to be tuned by minimizing again both the $MSE_u$ and $MSE_s$ on the region data. The pre-trained model is used to accelerate each region's training process.

### 3.Result

In this section, we conduct extensive evaluations of the proposed method. COVID-19 heatmap snapshots for the first 120 days (120 snapshots) are collected. The snapshot matrix assembles the information regarding 106 days for training, while DR-RNNs approximates the results for 120 days. The idea is to forecast two weeks in the future, given the data observed in the past 106 days. We present MSE between the real data and the DR-RNNs prediction for the 120$^{th}$ day in Figure.4. Table.1 and Figure.4 show the overall MSE for the compartments (susceptible, exposed, infectious, recovered, and deceased) approximations and computation time.

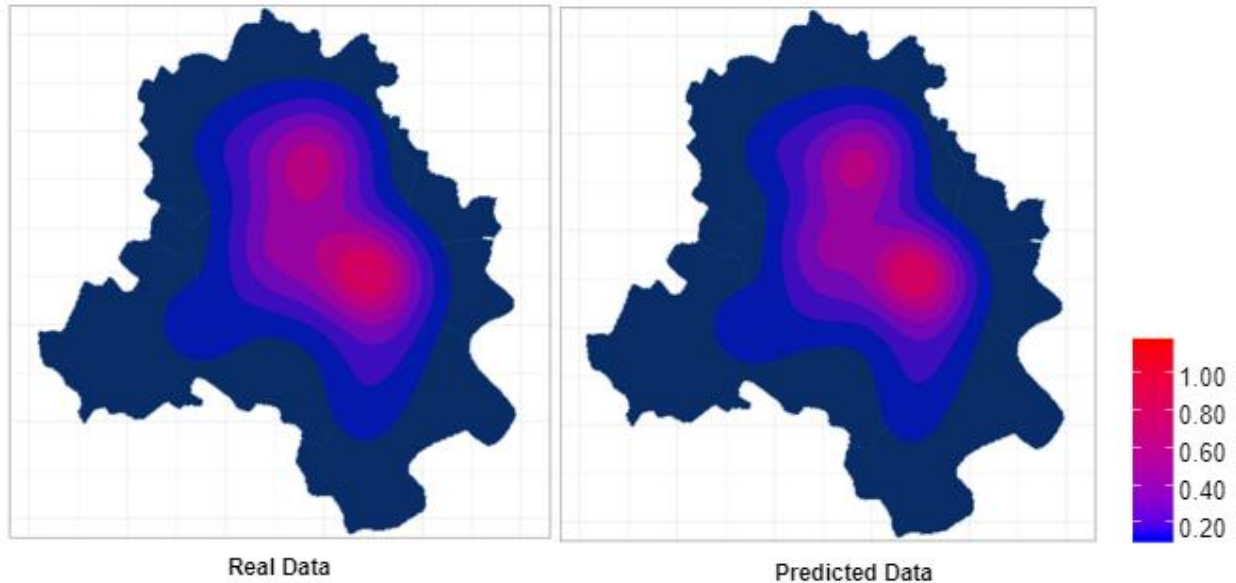

Figure 4: Comparison between a snapshot of real data and DR-RNNs prediction at $t = 120$ days. The DR-RNNs correctly captures the virus dynamics and accurately reproduces the solution with an MSE of $2.79e^{-3}$ in space and time.

Table.1. The MSE for the six compartment approximations and computation time

| Compartments | mean square error | Solution Time(seconds) |
|---|---|---|
| Susceptible | $1.54 \times 10^{-3}$ | 76.4 |
| Exposed | $2.74 \times 10^{-3}$ | 61.9 |
| Infectious | $1.30 \times 10^{-3}$ | 77.3 |
| Recovered | $2.54 \times 10^{-3}$ | 81.9 |
| Deceased | $1.94 \times 10^{-3}$ | 88.5 |

We can note that most compartments show results in agreement with the real data, while the exposed compartment reveals more pronounced differences than the other compartments, as shown in Table.1. This discrepancy occurs due to the different parameters for each equation in the SEIRD model, which largely affects the system's dynamics. The dynamics for each compartment are different since each compartment presents different coupling, diffusion, and reaction parameters.

## 4.Discussion and Limitations:

The curves for each compartment are different, which is the first thing we find. This discrepancy occurs due to the different parameters for each equation in the SEIRD model, which predominantly affects the system's dynamics. The dynamics for each compartment are different since each compartment presents different coupling, diffusion, and reaction parameters. Also, regarding this issue, since the parameters are time and space-dependent, sudden changes in their values can affect the system's dynamics and DR-RNNs dynamics mapping ability. Some sudden changes in the susceptible and exposed compartments related to stricter public policies considered to reduce the transmission rates (parameters infectious and exposed) are incorporated into the model. Since the parameter variation is not introduced smoothly, DR-RNNs ability to map sudden changes in the system's dynamics is reflected by some spikes on the mean square errors in time curves. Comparing the real data and prediction data, we observe that the errors tend to grow as soon as the forecasting starts for the 118$^{th}$ day. The exposed compartment, which yielded most of the oscillations due to parameter changing on the real data, presented the same behavior on the prediction phase around the 118$^{th}$ day. We also note that the exposed compartment yields a significant mean square error for the 120th day than the other compartments. In Figures.7, we compared these results with the presented results., we can conclude that the predictions are reasonably accurate compared to the real data, especially when considering the time required for calculation. In this study, the total population during the simulation, normalized by the total population modeled in the initial conditions. The DR-RNNs model does not consider population growth, and the value must be theoretically constant for all the simulations.

## 5.Conclusion

For public health agencies to effectively and timely plan hospital treatment and other resources available to end the outbreak, accurate COVID-19 case forecasting is a significant concern. We introduced DR-RNNs, a data-driven deep learning approach based on physics informed neural network to solve the SIERD model's daily time-varying parameters. The study's results will aid policymakers and healthcare providers in effectively prepare and provide resources to deal with the crisis in the coming days and weeks, including nurses, beds, and intensive care units. It is shown that the predictions made by DR-RNNs align well with the real data. Although the DR-RNNs can still predict the responses with sufficient accuracy, limited training data does not significantly affect the prediction performance of the physics-based learning method. The data should be updated in real-time for more precise comparisons and future perspectives.